# Machine and Deep Learning for Crowd Analytics

M. Siraj

**Abstract:** In high population cities, the gatherings of large crowds in public places and public areas accelerate or jeopardize people safety and transportation, which is a key challenge to the researchers. Although much research has been carried out on crowd analytics, many of existing methods are problem-specific, i.e., methods learned from a specific scene cannot be properly adopted to other videos. Therefore, this presents weakness and the discovery of these researches, since additional training samples have to be found from diverse videos. This paper will investigate diverse scene crowd analytics with traditional and deep learning models. We will also consider pros and cons of these approaches. However, once general deep methods are investigated from large datasets, they can be consider to investigate different crowd videos and images. Therefore, it would be able to cope with the problem including to not limited to crowd density estimation, crowd people counting, and crowd event recognition. Deep learning models and approaches are required to have large datasets for training and testing. Many datasets are collected taking into account many different and various problems related to building crowd datasets, including manual annotations and increasing diversity of videos and images. In this paper, we will also propose many models of deep neural networks and training approaches to learn the feature modeling for crowd analytics.

## 1. Introduction

The important increase in CCTV videos in the lastdecade has led to more video data being collected than can be investigated by a computer operator [1][2]. Indeed, vision based and real-time computations of these significantly increasing databases has become important issue for the machine learning and computer vision researchers [3][4][5]. The performance considering real-time computational overhead is very significant in large-data driven scenes where scalability and a rapid response time are required [6][7].

Montgomery presented a report where he mentioned that, more than half of the people of the world reside in dense cities [8][9][10]. In fact, automated crowd investigation plays an crucial role in crowd analysis and visual surveillance videos considering these CCTV cameras and other installation systems [11][12]. Therefore, in terms of designing public spaces, visual surveillance systems, and intelligent controlled physical situations [13]. Therese kind of approaches will have various important applications such as the monitoring of crowd flows, taking care of accidents, and managing evacuation designs required in the bad event of a sudden and uncontrolled fire or in presence of riots in cities zones especially [14][15][16]. In the research documentation, the researchers have investigated the situation of gathering the motion information at a higher level [17][18]. It means that the motion information does not take into account individual moving or static objects [19][20[21]. These methods therefore, often need various features including multi-resolution histograms [22][23], spatio-temporal cuboids [24][25], appearance or motion descriptors [26][27] and spatio-temporal cubes [28][29].

For this purpose, we need to understand the combine distribution of the image pixels [30]. Furthermore, to take into account temporal along with the spatial information of the data, the combine characteristics of the pixels across multiple adjacent video frames must be investigated [31][32]. For understanding, analyzing and learning, we make the general postulate that the distribution does not matter, it could be stationary over the learning interval or it could be mobile [33]. To consider the validity of this approach, it may be required to limit the temporal and spatial length of the learning window or time interval and therefore the number of videos in the training samples [34][35]. Our method is to understand, absorb and learn the distribution for a definite frame. Once this concept is understood and learned, it can be effectively prolonged to larger frame chunk sizes depending either an AR (Markov), MA, or ARMA process model [37]. Decreasing or imposing conditions on the training session reduce the number of learned parameters and therefore, the order of the process, hence reducing the learning variance. For this purpose, the flow diagram is presented as

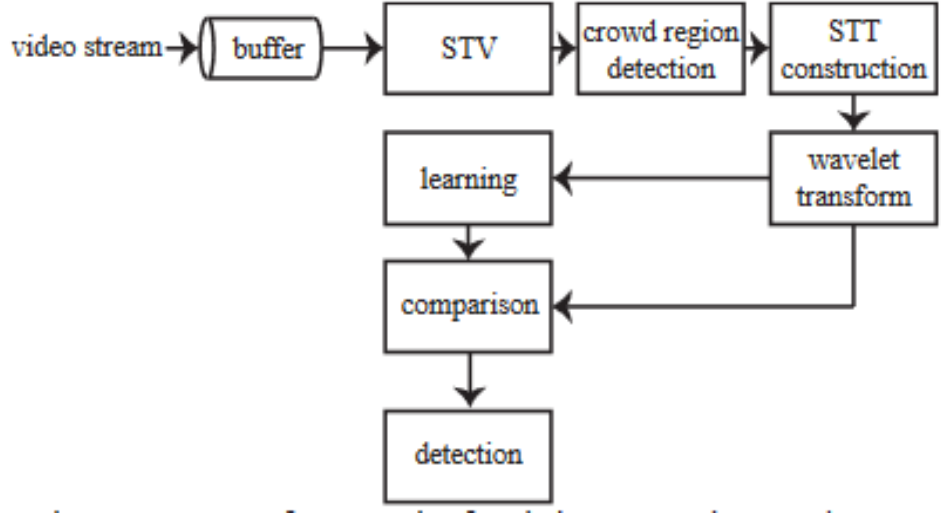

## 2. Proposed Method

We compute the estimation of the spatio-temporal pixel distribution, the traditional machine learning method [6] is to estimate whether or not a scene in the image or video is anomalous by computing its likelihood where we consider the learned distribution. This method is tending toward the fact that the anomalous situation is unknown, so the probability ratios cannot be properly computed. As an approach of how proper the method demonstrates the data, the Bayesian model is a significant feature. Thresholding the likelihood probability is encouraged by theoretical background considerations [6].

$$\ell(x) = (x - \mu_x)^T \Sigma^{-1} (x - \mu_x)$$

For each parameter, the partial derivative of the crowd model is computed for single training sequences of the videos under observations, i.e., one weight for each feature function Fj is considered. The partial derivative with respect to the parameter of the learning model corresponds to important value of the feature function for its true parameter, minus the averaged values of the feature function for all possible cases. Therefore, Eq. can be formulated as:

$$\frac{\partial}{\partial w_j} \log p(y/\mathbf{x}; \mathbf{w}) = F_j(\mathbf{x}, y) - \sum_{y'} p(y'/\mathbf{x}; \mathbf{w}) [F_j(\mathbf{x}, y')]$$

During the model development, it has been found that with fixed values, the odel approximately consider Gaussian distribution. This experimental estimation works properly on many testing videos for anomaly detection, and has been exploited for developing the normal crowd activities in this research. For each learning video, the formulation is defined as

$$\mu_{jk} = \frac{1}{L}\Sigma_i f_{ijk},$$

$$\sigma_{jk} = \sqrt{\frac{1}{L}\Sigma_i(f_{ijk} - \mu_{jk})^2}.$$

Due to the variety of moving individuals in a crowd video, well organized tracking individual objects is challenging. We demonstrate the crowd motion by the patch-based local motion formulation. Similarly to [6], the non-moving elements in the video are formulated as a collection of spatio-temporal cubes of dimension equal to p×p×q, where p(spatial size) and q(temporal size) must be big enough to encode the important characteristics of the different elements of the local motion flow. Every block is analyzed by a dynamic texture model [10], which is in fact a linear dynamic system of different parameters as formulated in the eq.

$$\begin{aligned} x_{t+1} &= Ax_t + Bv_t \\ y_t &= Cx_t + w_t, \end{aligned}$$

The same concept has been predicted in the Figure as shown below.

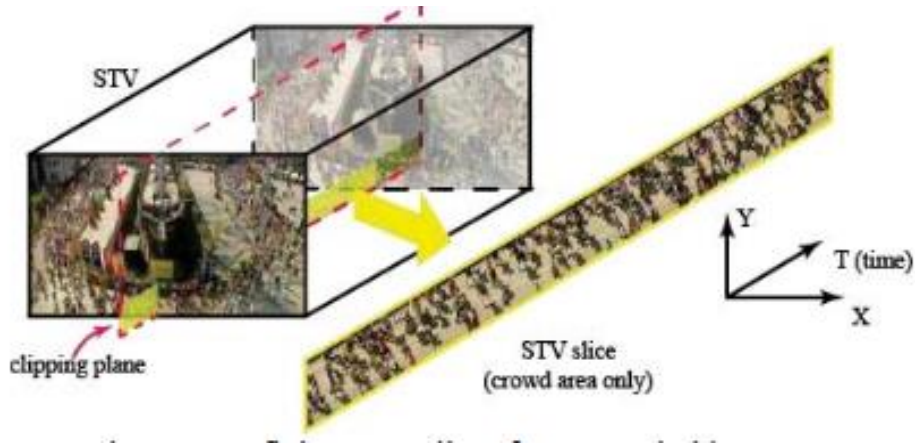

The important parameters for a motion pattern to be considered and learned are mean vector and covariance matrix specific to crowd scene under observation. Assume we have a motion pattern computed by m samples, if we get a new motion pattern whose training number is n, we estimate the new mean vector and covariance matrix of the new motion pattern as formulated:

$$\mu_c = \frac{m\mu_a}{m+n} + \frac{n\mu_b}{m+n}$$

$$\Sigma_c \approx \Sigma_c' = \frac{(m-1)\Sigma_a + n\Sigma_b}{m+n-1}$$

The same concept of the formulation of the above equation is highlighted in the figure below.

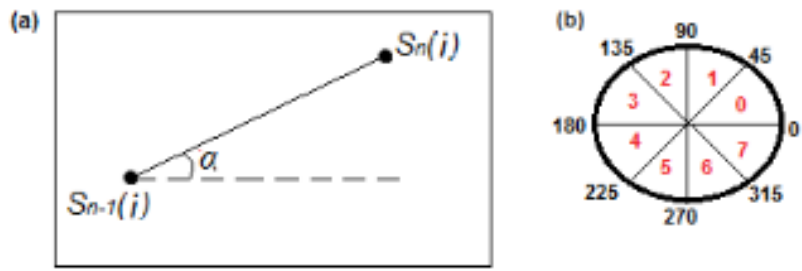

## 3. Experimental Analysis and Evaluations

In this section we discuss about the experimental analysis and evaluations, results, evaluation and performance of our proposed method. Our proposed method is implemented using MATLAB by modeling user interface. This method is implemented considering image processing libraries and tested on the dataset from University of Minnesota. For this purpose, we first convert videos into frames. These videos have normal and abnormal situations occurrences at different parts of the videos. In our experiment we have used the GROUND sequence for performance analysis. In the figure given below we show the results of our method obtained with the marathon video sequence. In the beginning, user interface is launched which consists of the input option for the video of the crowd. Secondly, each video frame is taken as input in a given window interval. As can be seen in the figure, important tracklets are highlighted in different colors to show these individuals in the crowd.

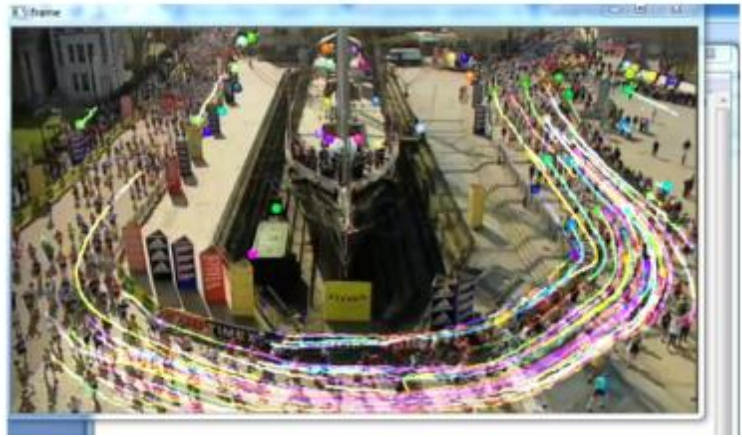

The Table shows the experimental analysis results in term of accuracy all video sequences. The results demonstrate that most of video sequences are accurately understood and learned by the algorithm. On average our proposed method achieved 88.83% accuracy when applied on the videos from the same dataset. Hence, it shows the effectiveness of our method.

| Runs | 1 | 2 | 3 | 4 | 5 |
|---|---|---|---|---|---|
| Accuracy | 91.67% | 93.33% | 93.33% | 100% | 93.33% |
| Runs | 6 | 7 | 8 | 9 | 10 |
| Accuracy | 100% | 100% | 60% | 63.33% | 93.33% |
| Average | 88.83% | | | | |

The same analysis are also performed in terms of different graphs. Both graphs below show that our method performs very correctly irrespective of the challenge of the crowd events.

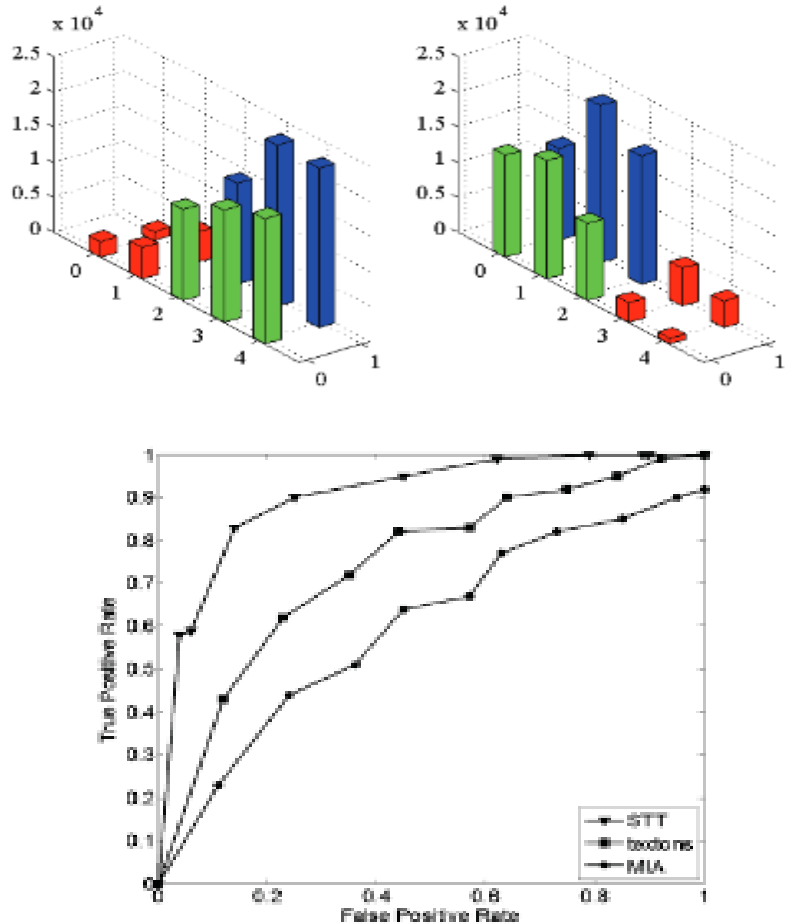